\documentclass{article}

\usepackage{arxiv}

\usepackage[utf8]{inputenc} % allow utf-8 input
\usepackage[T1]{fontenc}    % use 8-bit T1 fonts
\usepackage{hyperref}       % hyperlinks
\usepackage{url}            % simple URL typesetting
\usepackage{booktabs}       % professional-quality tables
\usepackage{amsfonts}       % blackboard math symbols
\usepackage{nicefrac}       % compact symbols for 1/2, etc.
\usepackage{microtype}      % microtypography
\usepackage{lipsum}
\usepackage{graphicx}

\usepackage{url}
\usepackage{xcolor}

\usepackage{hyperref}

\usepackage{makecell}
\usepackage{longtable}
\usepackage{tablefootnote}

\usepackage{amssymb}
\usepackage{latexsym}
\usepackage{threeparttable}
\usepackage{tikz}
\usetikzlibrary{arrows.meta,positioning,fit,shapes.misc}

\usepackage{subcaption}
\usepackage{tabularx}
\usepackage{booktabs} % for better table lines
\usepackage{array}
\usepackage{float}
\usepackage{siunitx}
\usepackage{lmodern}
\usepackage{multirow}
\usepackage{amsmath}
\usepackage{xcolor} % for row coloring commands
\usepackage{here}
\usepackage{colortbl}
\usepackage{natbib}
\title{CornOrb: A Multimodal Dataset of Orbscan Corneal Topography and Clinical Annotations for Keratoconus Detection}

\author{
Mohammed El Amine Lazouni \\
Biomedical Engineering Laboratory \\
Abou Bakr Belkaid University \\
Tlemcen, Algeria \\
\And
Leila Ryma Lazouni \\
Biomedical Engineering Laboratory \\
Abou Bakr Belkaid University \\
Tlemcen, Algeria \\
\And
Zineb Aziza Elaouaber \\
Biomedical Engineering Laboratory \\
Abou Bakr Belkaid University \\
Tlemcen, Algeria \\
\And
Mohammed Ammar \\
LIST Laboratory \\
M'Hamed Bougara Boumerdes University \\
Algeria \\
\And
Sofiane Zehar \\
Lazouni Clinic \\
Tlemcen, Algeria \\
\And
Mohammed Youcef Bouayad Agha \\
Lazouni Clinic \\
Tlemcen, Algeria \\
\And
Ahmed Lazouni \\
Lazouni Clinic \\
Tlemcen, Algeria \\
\And
Amel Feroui \\
Biomedical Engineering Laboratory \\
Abou Bakr Belkaid University \\
Tlemcen, Algeria \\
\And
Ali H. Al-Timemy \\
College of Excellence \\
University of Baghdad \\
Iraq \\
\And
Siamak Yousefi \\
Bascom Palmer Eye Institute \\
University of Miami \\
USA \\
\And
Mostafa El Habib Daho\thanks{Corresponding author: mostafa.elhabibdaho@univ-brest.fr} \\
University of Western Brittany \\
LaTIM UMR1101, Inserm \\
Brest, France \\
}

\begin{document}
\maketitle
\begin{abstract}
In this paper, we present CornOrb, a publicly accessible multimodal dataset of Bausch \& Lomb Orbscan 3 corneal topography images and clinical annotations collected from patients in Algeria. 

The dataset comprises 1,454 eyes from 744 patients, including 889 normal eyes and 565 keratoconus cases. 
For each eye, four corneal maps are provided (axial curvature, anterior elevation, posterior elevation, and pachymetry), together with structured tabular data including demographic information and key clinical parameters such as astigmatism, maximum keratometry (K\textsubscript{max}), central and thinnest pachymetry, and anterior/posterior asphericity. 

All data were retrospectively acquired, fully anonymized, and pre-processed into standardized PNG and CSV formats to ensure direct usability for artificial intelligence research. 
This dataset represents one of the first large-scale Orbscan 3-based resources from Africa, specifically built to enable robust AI-driven detection and analysis of keratoconus using multimodal data. 
The data are openly available at Zenodo (\href{https://doi.org/10.5281/zenodo.17127265}{10.5281/zenodo.17127265}).
\end{abstract}

% keywords can be removed

% -------------------------------
\section*{Specifications Table}
\begin{tabular}{p{4cm} p{10cm}}
\hline
\textbf{Subject area} & Ophthalmology, Medical Imaging, Artificial Intelligence \\

\textbf{More specific subject area} & Keratoconus detection and analysis using corneal topography and multimodal clinical data \\

\textbf{Type of data} & PNG images (axial curvature, anterior elevation, posterior elevation, pachymetry maps); CSV file (demographic and clinical annotations) \\

\textbf{How the data were acquired} & Routine clinical examinations performed with the Orbscan~3 device; exported from PDF reports, anonymized, and structured into standardized image and CSV formats \\

\textbf{Data format} & Raw data: PDF exports (not released); Processed data: anonymized PNG images and structured CSV metadata \\

\textbf{Parameters for data collection} & Retrospective data from 744 patients (1,454 eyes) collected in an Algerian ophthalmology center between 2017 and 2023; both normal and keratoconus cases included \\

\textbf{Description of data collection} & Each patient has a unique pseudonymized code. For each eye, four Orbscan 3 corneal maps (axial, anterior, posterior, pachymetry) were extracted and linked to clinical annotations (age, sex, astigmatism, maximum keratometry, central and thinnest pachymetry, anterior and posterior asphericity). Diagnostic labels (normal vs.~keratoconus) were assigned by clinicians \\

\textbf{Data accessibility} & Publicly available at Zenodo: \href{https://doi.org/10.5281/zenodo.17127265}{10.5281/zenodo.17127265} \\
\hline
\end{tabular}

% -------------------------------
\section*{Value of the Data}
\begin{itemize}
  \item The CornOrb dataset is one of the first large-scale, publicly accessible Orbscan 3 corneal topography resources from Africa, filling a critical gap in keratoconus research.
  \item By combining four standardized corneal maps with structured demographic and clinical annotations, it provides a comprehensive multimodal resource.
  \item Researchers in artificial intelligence (AI) can directly use the dataset to train, validate, and benchmark machine learning models for keratoconus detection, as well as explore multimodal learning strategies that integrate imaging and tabular data.
  \item Clinicians and biomedical engineers can use the dataset as a comparative reference to study Orbscan 3 measurements, support multi-center collaborations, and evaluate device-specific variability in corneal topography.
  \item The dataset supports reproducibility and transparency in ophthalmic AI by offering standardized, anonymized data in PNG and CSV formats, enabling fair benchmarking across algorithms and imaging modalities.

\end{itemize}

% -------------------------------
\section*{Objective}
The objective of this dataset article is to present CornOrb, a multimodal collection of Orbscan 3 corneal topography images and structured clinical annotations designed to support research on keratoconus. The dataset enables the development, validation, and benchmarking of AI models by combining four corneal maps with key clinical parameters. By releasing one of the first large-scale Orbscan 3-based resources, CornOrb addresses the lack of publicly accessible datasets in this domain and promotes reproducibility, transparency, and FAIR data use in ophthalmic AI research.

%%%%%%%%%%%%%%%%%%%%%%%%%%%%%%%%%%%%%%%%%%%%%%%%%%%%%%%%%%%%%%%%%%%%%%%%%%%%%%%%%%%%%%%%%%%%%%%%%
\section*{Background}
Keratoconus is a progressive corneal ectatic disorder characterized by localized thinning and protrusion of the cornea, leading to irregular astigmatism and gradual visual decline \citep{Rabinowitz1998,Godefrooij2017}. Early and reliable detection is essential to guide timely interventions and reduce the risk of advanced disease requiring corneal transplantation \citep{Ferdi2019}.  

Corneal topography plays a central role in clinical diagnosis, with maps providing complementary information on anterior and posterior corneal surfaces as well as pachymetric profiles \citep{Ghemame2019,Cairns2005}. These representations are particularly important for detecting early keratoconus and monitoring disease progression \citep{Lazouni2025ARVO,Lazouni2024ARVO,Hazarbassanov2025ARVO}.  

Despite the growing integration of AI into ophthalmology \citep{Tiong2024,Syta2025}, publicly available datasets for corneal diseases remain scarce, and large-scale resources based on Orbscan 3 imaging are virtually absent. In recent years, however, several authors have shared keratoconus datasets in open access. Bakir et al. released a Pentacam-based dataset for stage detection of keratoconus \citep{Bakir2023}. Al-Timemy et al. first introduced a large Pentacam dataset of 4,844 corneal maps, together with a hybrid deep learning model for keratoconus detection \citep{AlTimemy2021}, and later published a multi-center Pentacam collection for early detection across different populations \citep{AlTimemy2023}. Castro de Luna et al. provided multiple tabular datasets including Pentacam indices, aberrometry parameters, and visual limitation records \citep{CastroLuna2020,CastroLuna2019}. In addition, Yousefi et al. released a large dataset of tabular parameters derived from OCT-based tomography \citep{Yousefi2018}.

The CornOrb dataset was developed to address the persistent gap in large-scale Orbscan 3 data, offering a multimodal collection of topography images and clinical annotations to support AI-driven research and clinical studies on keratoconus.

% -------------------------------
\section*{Data Description}

\begin{figure}[t!]
\centering
\begin{tabular}{cc}
    \includegraphics[width=0.25\linewidth]{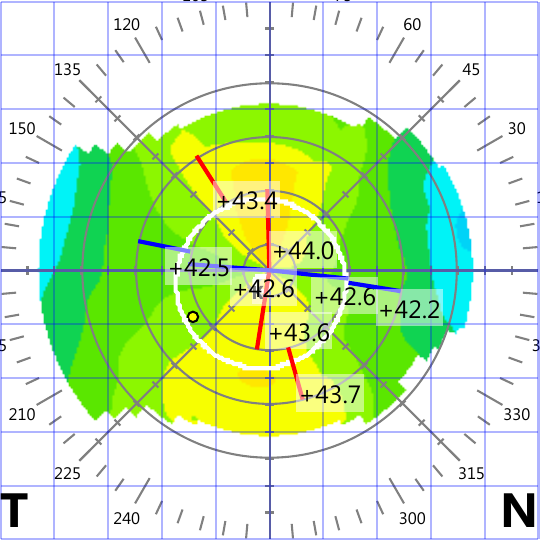} &
    \includegraphics[width=0.25\linewidth]{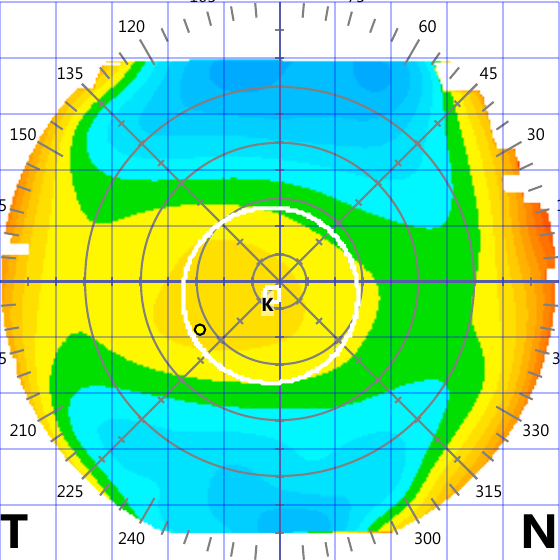} \\
    (a) Axial curvature & (b) Anterior elevation \\
    \includegraphics[width=0.25\linewidth]{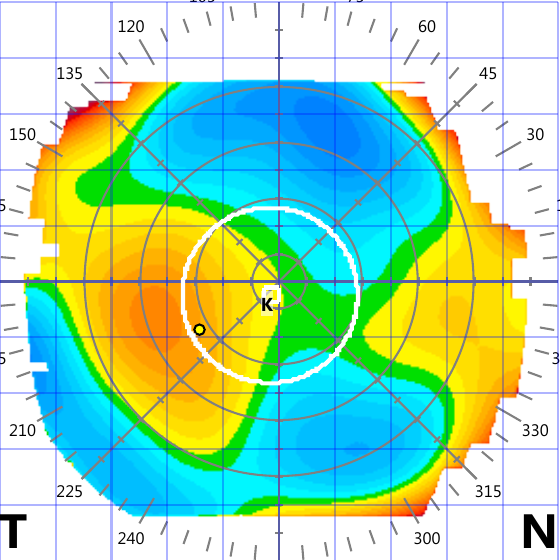} &
    \includegraphics[width=0.25\linewidth]{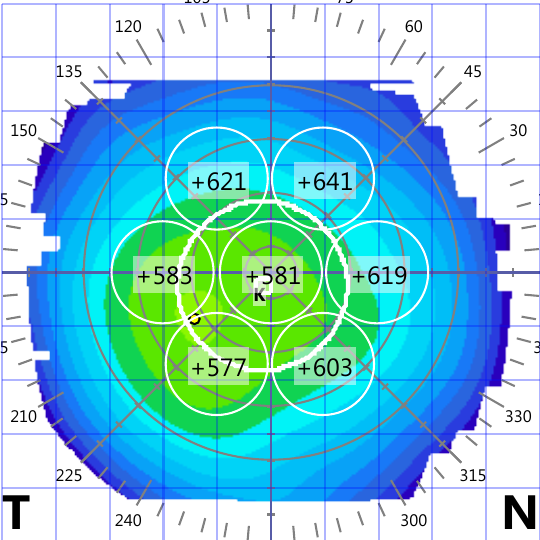} \\
    (c) Posterior elevation & (d) Pachymetry \\
\end{tabular}
\caption{Representative Orbscan~3 corneal maps included in the CornOrb dataset: 
(a) axial curvature, (b) anterior elevation, (c) posterior elevation, and (d) pachymetry. 
Each map is standardized to a resolution of $560 \times 560$ pixels and saved in PNG format.}
\label{fig:corneal-maps}
\end{figure}

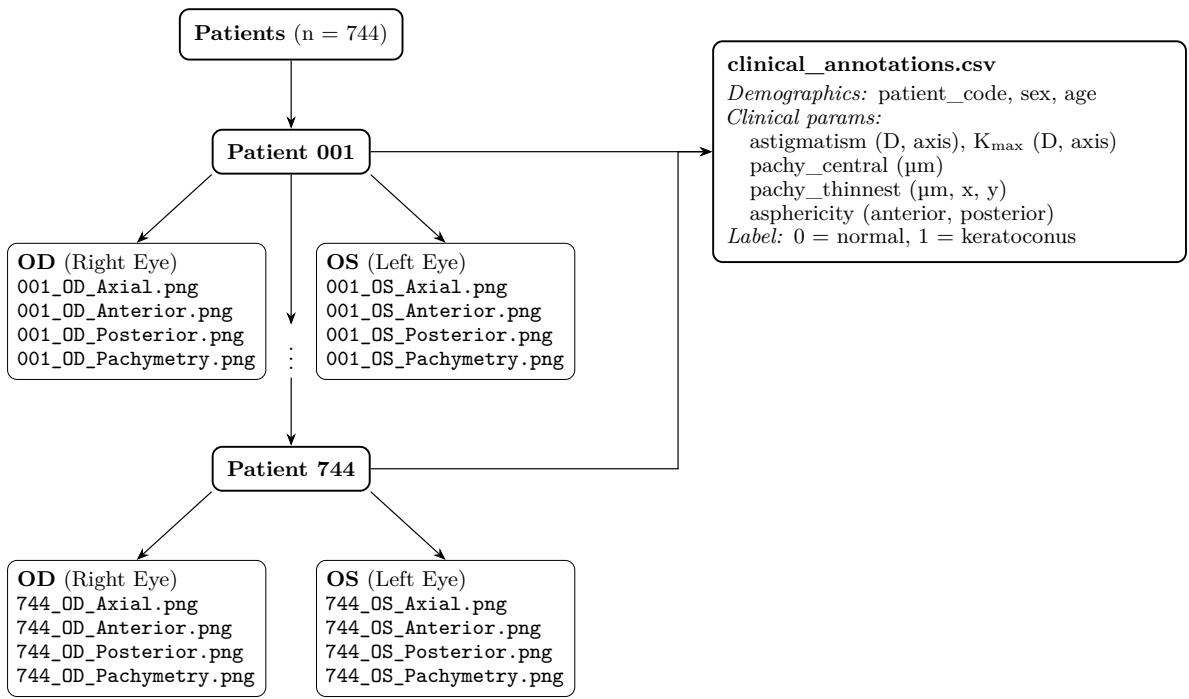
\begin{figure}[t!]
\centering
\resizebox{0.95\linewidth}{!}{%
\begin{tikzpicture}[
  font=\small,
  node distance=10mm and 12mm,
  box/.style={draw, rounded corners, thick, inner sep=6pt, align=left},
  thinbox/.style={draw, rounded corners, inner sep=4pt, align=left},
  >={Stealth[length=2.2mm]},
  arrow/.style={-Stealth, line width=0.5pt}
]

% Top: Patients
\node[box] (patients) {\textbf{Patients} (n = 744)};

% Patient 001
\node[box, below=of patients] (p1) {\textbf{Patient 001}};

% OD / OS under Patient 001
\node[thinbox, below left=10mm and -8mm of p1, text width=35mm] (p1od) {\textbf{OD} (Right Eye)\\
\texttt{001\_OD\_Axial.png}\\
\texttt{001\_OD\_Anterior.png}\\
\texttt{001\_OD\_Posterior.png}\\
\texttt{001\_OD\_Pachymetry.png}
};

\node[thinbox, below right=10mm and -8mm of p1, text width=35mm] (p1os) {\textbf{OS} (Left Eye)\\
\texttt{001\_OS\_Axial.png}\\
\texttt{001\_OS\_Anterior.png}\\
\texttt{001\_OS\_Posterior.png}\\
\texttt{001\_OS\_Pachymetry.png}
};

% Dots (...)
\node[below=22mm of p1] (dots) {$\vdots$};

% Patient 744
\node[box, below=of dots] (p744) {\textbf{Patient 744}};

% OD / OS under Patient 744
\node[thinbox, below left=10mm and -8mm of p744, text width=35mm] (p744od) {\textbf{OD} (Right Eye)\\
\texttt{744\_OD\_Axial.png}\\
\texttt{744\_OD\_Anterior.png}\\
\texttt{744\_OD\_Posterior.png}\\
\texttt{744\_OD\_Pachymetry.png}
};

\node[thinbox, below right=10mm and -8mm of p744, text width=35mm] (p744os) {\textbf{OS} (Left Eye)\\
\texttt{744\_OS\_Axial.png}\\
\texttt{744\_OS\_Anterior.png}\\
\texttt{744\_OS\_Posterior.png}\\
\texttt{744\_OS\_Pachymetry.png}
};

% CSV box on the right
\node[box, right=50mm of p1, text width=65mm, align=left] (csv) {\textbf{clinical\_annotations.csv}\\[2pt]
\textit{Demographics:} patient\_code, sex, age\\
\textit{Clinical params:}\\
\quad astigmatism (D, axis), K\textsubscript{max} (D, axis)\\
\quad pachy\_central (\textmu m)\\
\quad pachy\_thinnest (\textmu m, x, y)\\
\quad asphericity (anterior, posterior)\\
\textit{Label:} 0 = normal, 1 = keratoconus
};

% Arrows
\draw[arrow] (patients.south) -- (p1.north);
\draw[arrow] (p1.south west) -- (p1od.north);
\draw[arrow] (p1.south east) -- (p1os.north);
\draw[arrow] (p1.east) -- ++(15mm,0) -- (csv.west);

\draw[arrow] (p1.south) -- (dots.north);
\draw[arrow] (dots.south) -- (p744.north);

\draw[arrow] (p744.south west) -- (p744od.north);
\draw[arrow] (p744.south east) -- (p744os.north);
\draw[arrow] (p744.east) -- ++(45mm,0) |- (csv.west); % FIXED ARROW

\end{tikzpicture}%
}
\caption{Schematic representation of the CornOrb dataset structure. Data are organized hierarchically by patient (from 001 to 744) and eye (OD = right eye, OS = left eye). 
Each eye includes four Orbscan~3 corneal maps. 
A CSV file accompanies the images with demographic and clinical annotations. \textbf{Note:} D = diopters, the standard unit for refractive power and keratometry.}
\label{fig:dataset-structure}
\end{figure}

The CornOrb dataset comprises 1{,}454 eyes from 744 patients, of which 889 eyes are clinically normal and 565 are diagnosed with keratoconus. Each record corresponds to a complete Orbscan~3 examination and is organized hierarchically by patient code and eye laterality (OD for right eye, OS for left eye). Both eyes are included when available and may be analyzed independently, although potential inter-eye correlation should be considered. For each eye, four corneal maps are provided as anonymized PNG images with a fixed resolution of $560 \times 560$ pixels: axial curvature, anterior elevation, posterior elevation, and pachymetry (Fig.~\ref{fig:corneal-maps}). File names explicitly encode the patient identifier, eye laterality, and map type (e.g., \texttt{001\_OD\_Axial.png}), ensuring easy reproducibility and traceability in AI workflows. Fig.~\ref{fig:dataset-structure} summarizes the folder organization and links to the accompanying CSV file.

%%%%%%%%%%%%%%%%%%%%%%%%%%%%%%%%%%%%%%%%%%%%%%%%%%%%%%%%%%%%%%%%%%%%

%%%%%%%%%%%%%%%%%%%%%%%%%%%%%%%%%%%%%%%%%%%%%%%%%%%%%%%%%%%%%%%%%%%%%%%%%%%%%%%%%%%%%%%%%%%

%\newpage

In addition to the imaging component, a structured CSV file accompanies the dataset and links directly to the images. This file contains:
\begin{itemize}
    \item \textbf{Demographics:} patient code, sex, age.  
    \item \textbf{Clinical parameters:} astigmatism (diopters and axis), maximum keratometry (K\textsubscript{max}, diopters and axis), central pachymetry, thinnest pachymetry (value in $\mu$m and Cartesian coordinates), anterior/posterior asphericity.  
    \item \textbf{Diagnostic label:} binary indicator (0 = normal, 1 = keratoconus).  
\end{itemize}
No missing values are present in the clinical annotations. The detailed structure of the CSV file, including variable names, units, and descriptions, is provided in Table~\ref{tab:data-dictionary}.

\begin{table}[t!]
\centering
\caption{Data dictionary for the \texttt{clinical\_annotations.csv} file.}
\label{tab:data-dictionary}
\begin{tabular}{p{3.8cm} p{3.2cm} p{9cm}}
\hline
\textbf{Variable name} & \textbf{Unit / Type} & \textbf{Description} \\
\hline
\texttt{patient\_code} & string & Unique pseudonymized identifier assigned to each patient. \\
\texttt{sex} & categorical (M/F) & Patient sex. \\
\texttt{age} & years (integer) & Age at examination. \\
\texttt{eye} & categorical (OD/OS) & Eye laterality: OD = right eye, OS = left eye. \\
\texttt{astigmatism\_D} & diopters (float) & Magnitude of corneal astigmatism. \\
\texttt{astigmatism\_axis} & degrees (0--180) & Axis of astigmatism. \\
\texttt{kmax\_D} & diopters (float) & Maximum keratometry value (K\textsubscript{max}). \\
\texttt{kmax\_axis} & degrees (0--180) & Axis corresponding to K\textsubscript{max}. \\
\texttt{pachy\_central} & $\mu$m (integer) & Central corneal thickness. \\
\texttt{pachy\_thinnest} & $\mu$m (integer) & Thinnest corneal thickness. \\
\texttt{pachy\_thinnest\_x} & mm (float) & X coordinate of the thinnest corneal point, relative to the corneal apex. \\
\texttt{pachy\_thinnest\_y} & mm (float) & Y coordinate of the thinnest corneal point, relative to the corneal apex. \\

\texttt{asphericity\_anterior} & unitless\tablefootnote{Asphericity (Q-value) is a unitless parameter that quantifies the deviation of the corneal surface from a perfect sphere, and therefore has no physical units.} (float) & Anterior corneal asphericity (Q-value). \\
\texttt{asphericity\_posterior} & unitless (float) & Posterior corneal asphericity (Q-value). \\
\texttt{label} & binary (0/1) & Diagnostic label: 0 = normal, 1 = keratoconus. \\
\hline
\end{tabular}
\end{table}

%As summarized in Table~\ref{tab:summary}, keratoconus eyes show higher K\textsubscript{max}, thinner central and thinnest pachymetry, and more negative anterior/posterior asphericity compared with normal eyes.

The distribution of diagnostic labels exhibits a moderate class imbalance (889 normal vs. 565 keratoconus eyes), a typical characteristic of real-world clinical cohorts. This imbalance should be considered when benchmarking machine learning algorithms, for example, through stratified splitting, class weighting, or focal loss.

%\begin{figure}[h!]
%  \centering
%  \includegraphics[width=0.45\linewidth]{class_distribution.png}
%  \caption{Class distribution (n=1{,}454 eyes). The cohort includes 889 normal and 565 keratoconus cases.}
%  \label{fig:class-distribution}
%\end{figure}

%\newpage

Table~\ref{tab:summary} summarizes clinical parameters for normal and keratoconus eyes. 
Continuous variables are reported as mean $\pm$ SD. Between-group differences were assessed 
with two-sided Welch's $t$-tests, and $p$-values were adjusted for multiple comparisons 
using the Benjamini--Hochberg procedure to control the false discovery rate at 5\%; 
adjusted values are reported as $p_{\text{BH}}$. All variables differed significantly 
after correction (all $p_{\text{BH}}<0.001$).

\begin{table}[h!]
\centering
\begin{threeparttable}
\caption{Summary statistics with one additional column reporting the BH-adjusted $p$-value for the Normal vs.\ Keratoconus comparison.}
\label{tab:summary}
\begin{tabular}{lccc}
\toprule
\textbf{Variable} & \textbf{Normal (n=889)} & \textbf{Keratoconus (n=565)} & \textbf{$p_{\text{BH}}$} \\
\midrule
Age (years)                   & $33.5 \pm 7.4$   & $27.9 \pm 7.1$   & $<0.001$ \\
Central pachymetry ($\mu$m)  & $530 \pm 43$     & $465 \pm 54$     & $<0.001$ \\
Thinnest pachymetry ($\mu$m) & $521 \pm 44$     & $427 \pm 59$     & $<0.001$ \\
K\textsubscript{max} (D)     & $43.7 \pm 2.9$   & $53.1 \pm 6.9$   & $<0.001$ \\
Astigmatism (D)              & $-1.35 \pm 0.91$ & $-5.32 \pm 3.32$ & $<0.001$ \\
Anterior asphericity         & $-0.15 \pm 0.45$ & $-1.20 \pm 0.81$ & $<0.001$ \\
Posterior asphericity        & $-0.17 \pm 0.43$ & $-1.16 \pm 0.83$ & $<0.001$ \\
\bottomrule
\end{tabular}

\end{threeparttable}
\end{table}

\begin{figure}[t!]
\centering
\begin{subfigure}{0.32\linewidth}\includegraphics[width=\linewidth]{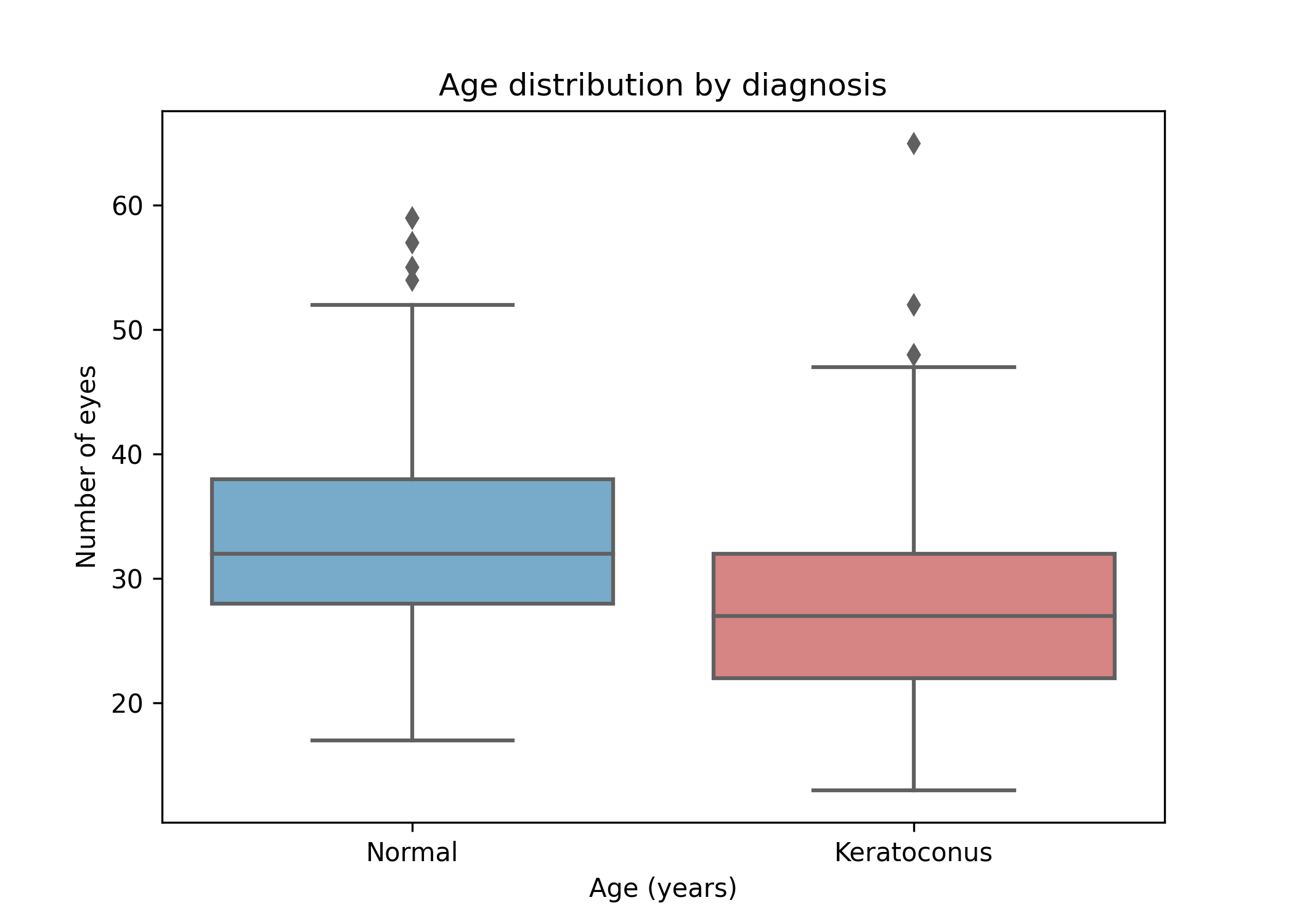}\caption{Age}\end{subfigure}
\begin{subfigure}{0.32\linewidth}\includegraphics[width=\linewidth]{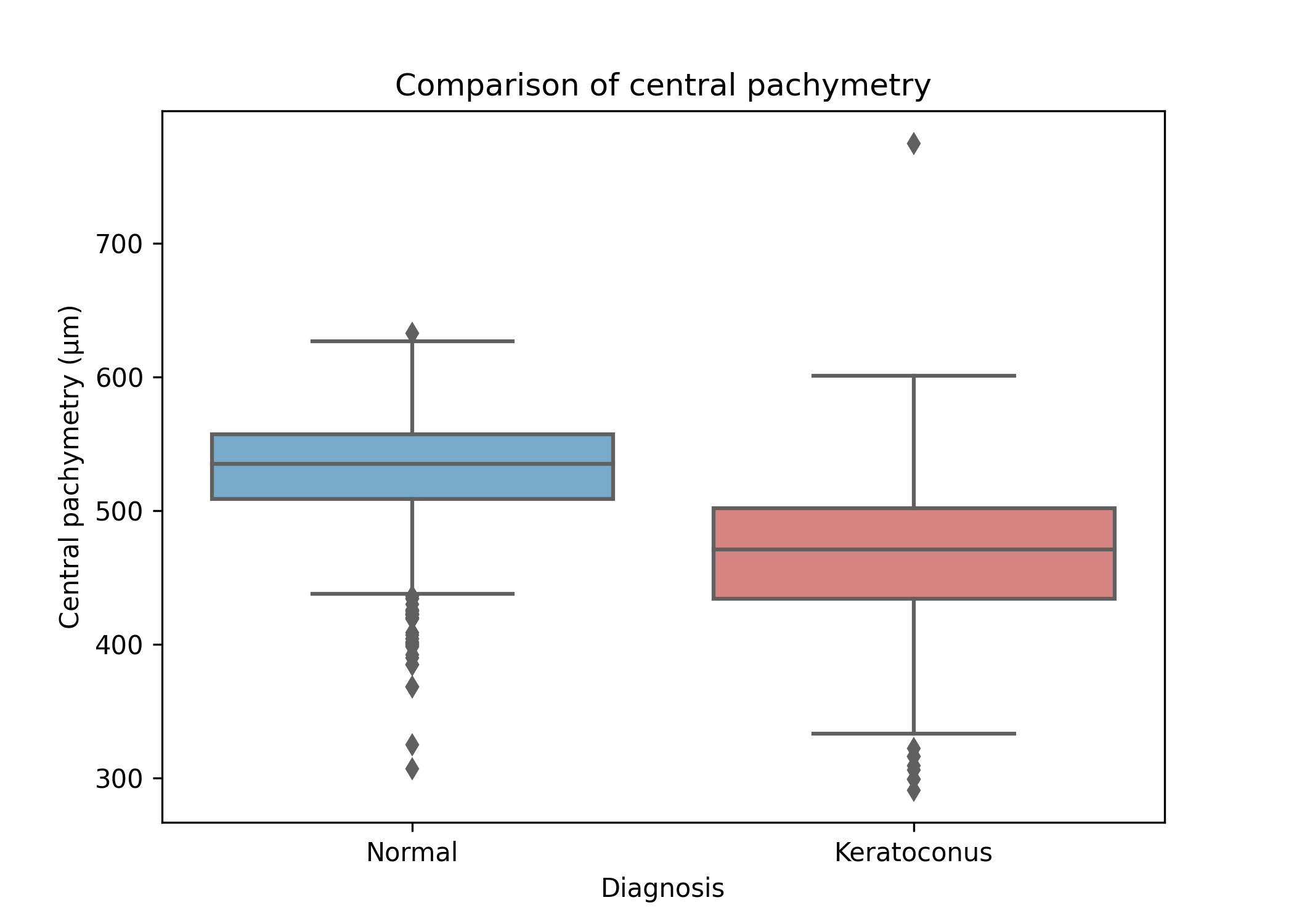}\caption{Central pachymetry}\end{subfigure}
\begin{subfigure}{0.32\linewidth}\includegraphics[width=\linewidth]{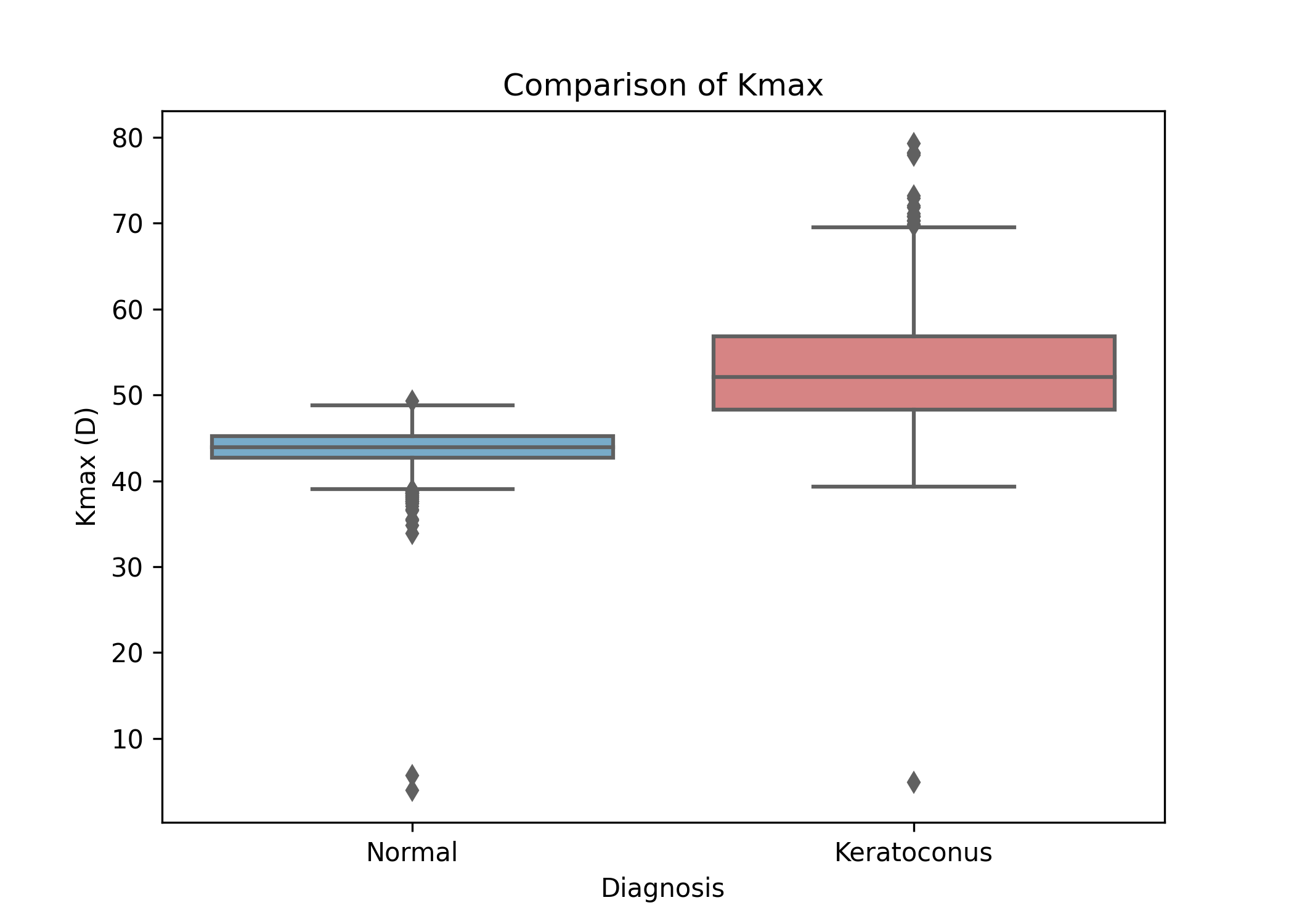}\caption{K\textsubscript{max}}\end{subfigure}

\begin{subfigure}{0.32\linewidth}\includegraphics[width=\linewidth]{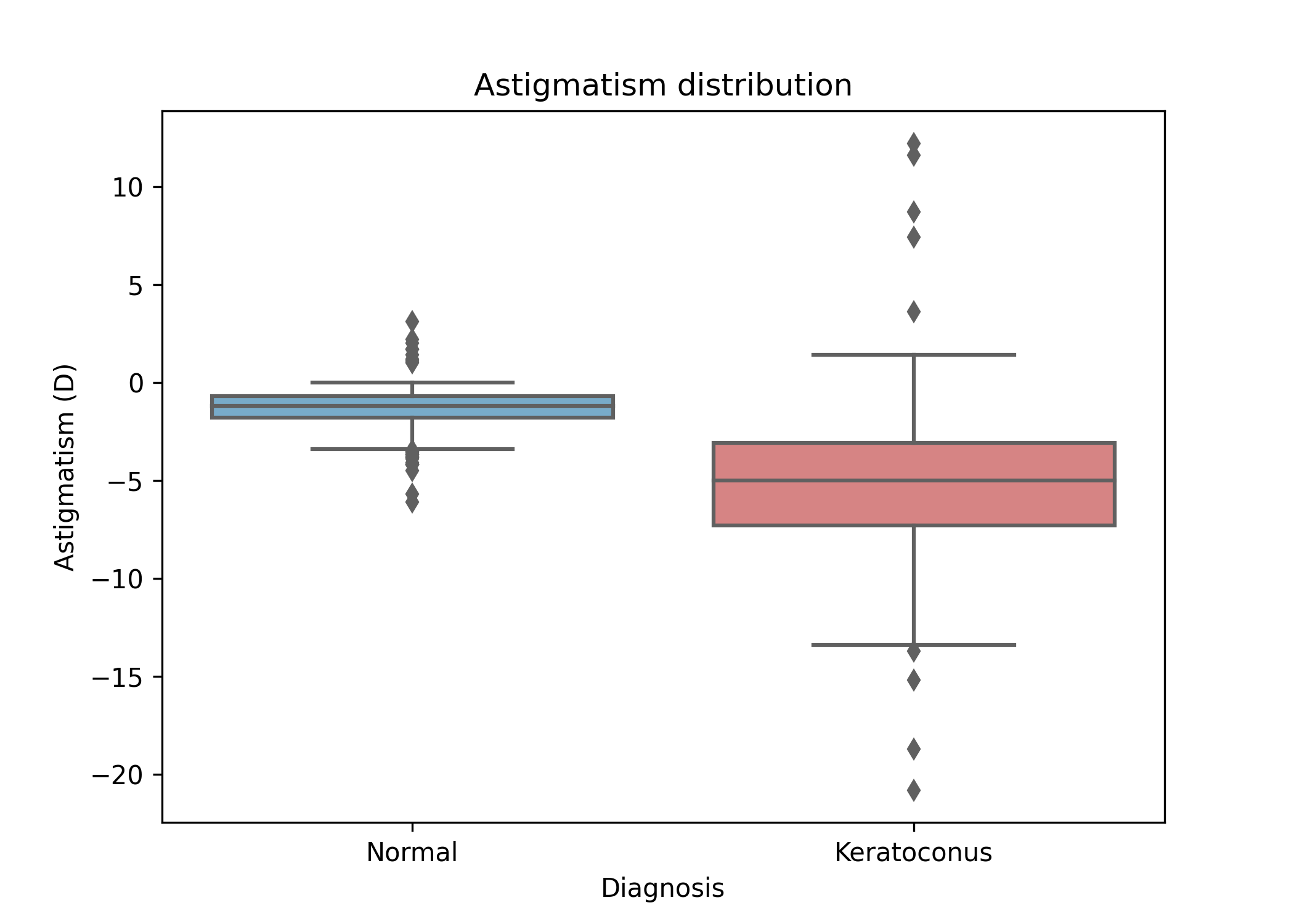}\caption{Astigmatism}\end{subfigure}
\begin{subfigure}{0.32\linewidth}\includegraphics[width=\linewidth]{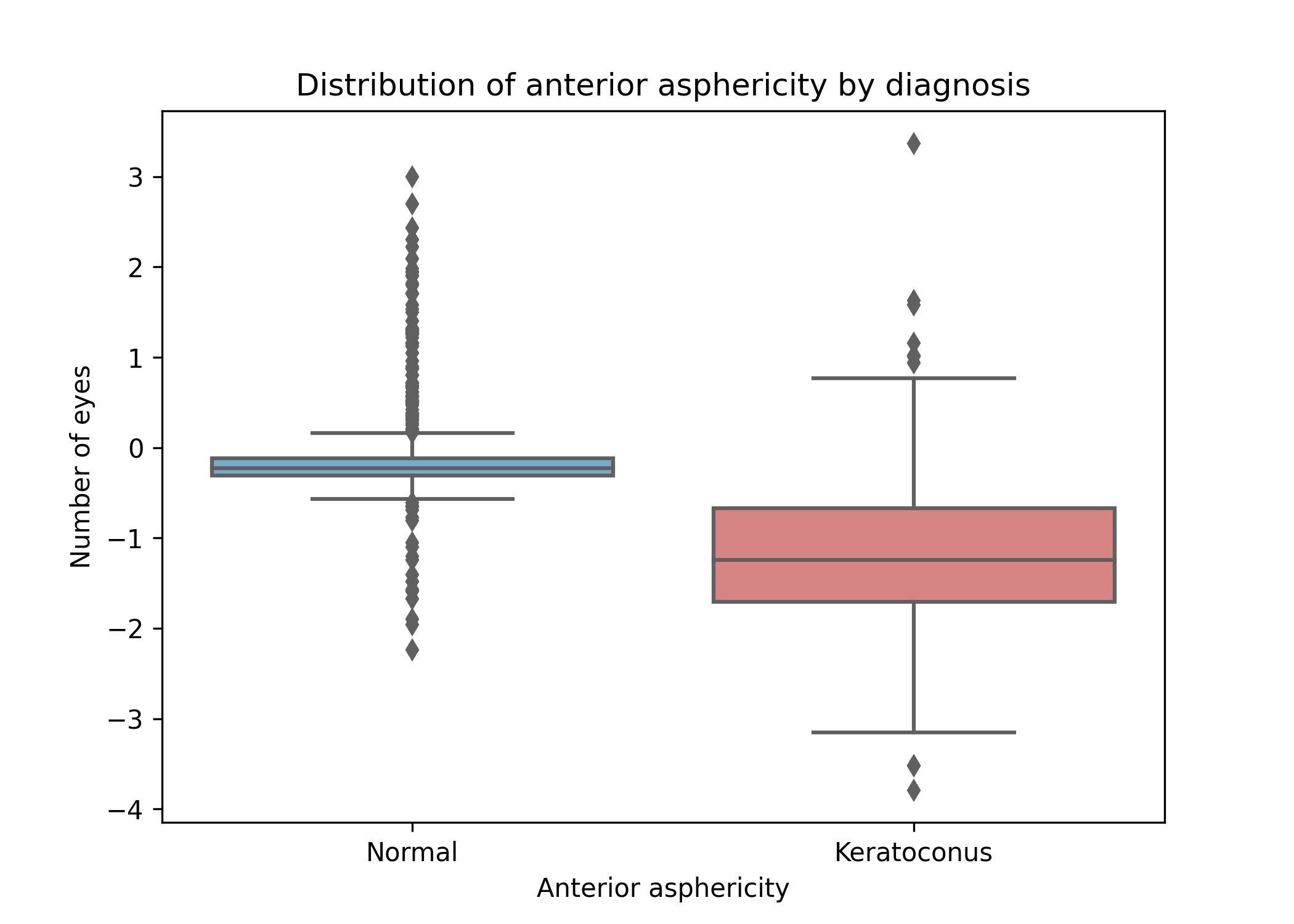}\caption{Anterior asphericity}\end{subfigure}
\begin{subfigure}{0.32\linewidth}\includegraphics[width=\linewidth]{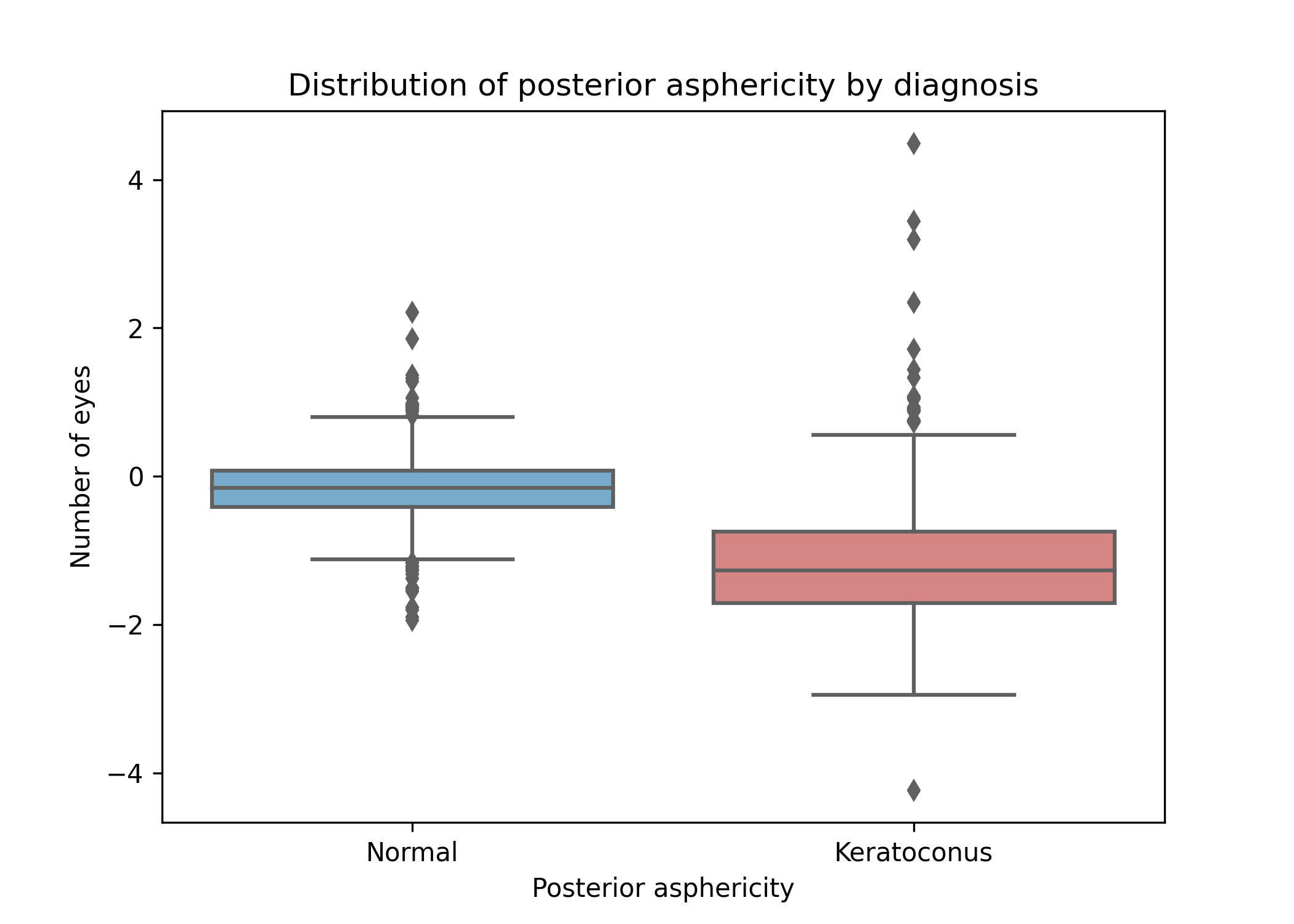}\caption{Posterior asphericity}\end{subfigure}
\caption{Distributions of quantitative clinical measures by diagnosis (Normal vs.\ Keratoconus). Panels display age (years), K\textsubscript{max} (D), central pachymetry ($\mu$m), astigmatism (D), and anterior/posterior asphericity (dimensionless). Colors: Normal (blue), Keratoconus (red).}

\label{fig:merged-box}
\end{figure}

\paragraph{Age distribution}
Patients with keratoconus are younger than those in the normal group (Table~\ref{tab:summary}; mean $27.9$ vs. $33.5$ years), which reflects the natural history of the disease and supports the inclusion of age as a covariate in predictive models (Fig.~\ref{fig:merged-box}(a)).

\paragraph{Central pachymetry}
Keratoconus eyes exhibit markedly reduced central corneal thickness compared to normal eyes ($465\,\mu$m vs. $530\,\mu$m), consistent with ectatic thinning (Fig.~\ref{fig:merged-box}(b)). This parameter is a key biomarker for clinical grading and staging.

\paragraph{Maximum keratometry (K\textsubscript{max})}
K\textsubscript{max} values are substantially higher in keratoconus eyes ($53.1$~D vs. $43.7$~D), capturing the characteristic steepening of the CornOrb (Fig.~\ref{fig:merged-box}(c)). This metric is widely used in both clinical and research settings as a diagnostic and prognostic indicator.

\paragraph{Astigmatism}
As shown in Fig.~\ref{fig:merged-box}(d), the magnitude of astigmatism displays a broader distribution and more negative values in keratoconus patients ($-5.32 \pm 3.32$~D) compared to normal eyes ($-1.35 \pm 0.91$~D). This irregular optical distortion is a hallmark of keratoconus and provides complementary discriminative information alongside curvature and pachymetry indices.

\paragraph{Asphericity (anterior and posterior)}  
Asphericity quantifies the deviation of the corneal surface from a perfect sphere. In normal eyes, both anterior and posterior asphericity values are close to zero, reflecting a mildly prolate corneal geometry (Fig.~\ref{fig:merged-box}(e)(f)). In keratoconus eyes, distributions shift markedly toward more negative values, consistent with corneal steepening and ectatic changes (Table~\ref{tab:summary}). This makes asphericity a valuable discriminative index for keratoconus detection and staging.

Taken together, the dataset provides a comprehensive representation of keratoconus through both image-based and quantitative descriptors. Its multimodal structure makes it suitable for supervised learning (classification), fair benchmarking, and multimodal fusion (image and tabular data).

% -------------------------------
\section*{Experimental Design, Materials and Methods}

\begin{figure}[t!]
  \centering
  \includegraphics[width=0.6\linewidth]{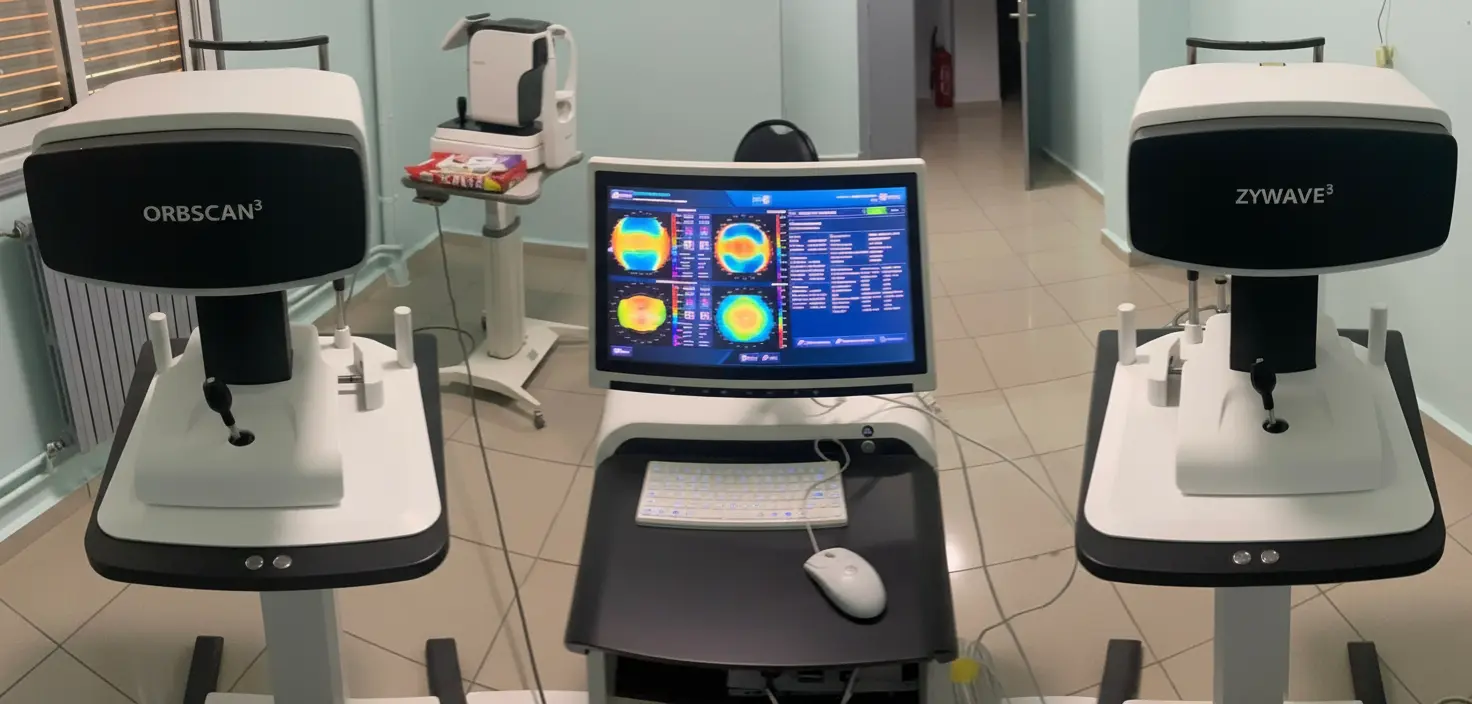}
  \caption{The Bausch \& Lomb Orbscan~3 device used for all corneal topography acquisitions in the CornOrb dataset.}
  \label{fig:orbscan}
\end{figure}

\paragraph{Acquisition setting}
All examinations were performed at Lazouni Clinic, an ophthalmology center in Tlemcen, Algeria, using a Bausch \& Lomb Orbscan~3 corneal topographer (Fig.~\ref{fig:orbscan}). Data were collected retrospectively from consecutive patients examined between 2017 and 2023. The cohort includes both clinically normal eyes and eyes diagnosed with keratoconus.  

%\paragraph{Ethics and consent}
%The study protocol adhered to the principles of the Declaration of Helsinki. All patients provided written informed consent for the secondary use of their anonymized data in research.  

\paragraph{Data export and preprocessing}
Raw examinations were initially exported as PDF reports generated by the Orbscan~3 device. A custom Python pipeline was developed to automatically parse the reports and extract the four corneal maps. Each map was cropped, standardized to $560 \times 560$ pixels, and saved in PNG format. All images were visually inspected, and poor-quality scans or incomplete examinations were excluded.  

\paragraph{Clinical annotation and labeling}
Structured tables embedded in the PDF reports were parsed to extract demographic and clinical parameters, which were compiled into a single CSV file. The annotations include sex, age, astigmatism magnitude and axis, maximum keratometry (K\textsubscript{max} and axis), central and thinnest pachymetry (values and coordinates), and anterior/posterior asphericity. Diagnostic labels were independently assigned by two junior ophthalmologists; in cases of disagreement, a senior ophthalmologist adjudicated the final label.  

\paragraph{Curation and organization}
All identifiers (e.g., patient names, record numbers, birth dates) were removed before release. Unique pseudonymized codes were assigned to each patient, with eye laterality encoded as OD (right) or OS (left). Filenames follow a standardized convention (\texttt{patientID\_eye\_maptype.png}), ensuring reproducibility and ease of automated processing. The CSV file links each eye to its corresponding maps.  

\paragraph{Quality control}
The dataset underwent several validation steps: (i) verification of consistent image resolution, (ii) cross-checking filenames against metadata, (iii) removal of duplicate or incomplete cases, and (iv) review of clinical variables for range validity. No missing values remain in the final release.  

\paragraph{FAIR compliance}
To promote reuse, the dataset was curated in line with the FAIR principles (Findable, Accessible, Interoperable, Reusable). Standard file formats (PNG for images, CSV for metadata) were used. Metadata fields were harmonized with ophthalmic AI conventions, and the dataset was deposited in an open-access repository (Zenodo) with a persistent DOI. 

% -------------------------------

\section{Limitations}

All imaging data were acquired using the Orbscan~3 device. This may limit generalizability to datasets collected with other corneal topographers (e.g., Pentacam, Galilei, Casia), since device-specific differences in curvature, elevation, or pachymetry measurements can affect model performance.  

The dataset represents a North African population from a single clinical center. While this provides a unique and underrepresented cohort, the geographic and ethnic homogeneity may restrict generalizability to populations with different corneal morphologies or risk factors for keratoconus.  

The dataset is retrospective and cross-sectional. It does not include longitudinal follow-up or explicit severity grading of keratoconus, which may limit studies on disease progression and staging.

Finally, the dataset also presents a moderate class imbalance (889 normal vs.~565 keratoconus eyes), reflecting real-world prevalence but requiring appropriate handling (e.g., stratified splitting, class weighting, or focal loss) when training AI models.

% -------------------------------
\section*{Ethics Statement}
This study was conducted in accordance with the principles of the Declaration of Helsinki. All patients provided written informed consent for the use of their anonymized clinical data in research. Data were fully anonymized prior to analysis and publication.

% -------------------------------
\section*{CRediT Author Statement}
\noindent \textbf{Mohammed El Amine Lazouni}: Methodology; Data Management; Writing -- Original Draft. \\
\textbf{Leila Ryma Lazouni}: Investigation; Software; Writing -- Review \& Editing. \\
\textbf{Zineb Aziza Elaouaber}: Investigation; Visualization; Writing -- Review \& Editing. \\
\textbf{Mohammed Ammar}: Formal Analysis; Software; Writing -- Review \& Editing. \\
\textbf{Sofiane Zehar}: Data Curation; Clinical Investigation; Writing -- Review \& Editing. \\
\textbf{Mohammed Youcef Bouayad Agha}: Data Curation; Clinical Investigation; Writing -- Review \& Editing. \\
\textbf{Ahmed Lazouni}: Resources; Data Curation; Clinical Investigation; Writing -- Review \& Editing. \\
\textbf{Amel Feroui}: Formal Analysis; Validation; Writing -- Review \& Editing. \\
\textbf{Ali Al-Timemy}: Formal Analysis; Validation; Writing -- Review \& Editing. \\
\textbf{Siamak Yousefi}: Supervision; Validation; Writing -- Review \& Editing. \\
\textbf{Mostafa El Habib Daho}: Conceptualization; Project Administration; Supervision; Writing -- Review \& Editing.

% -------------------------------
\section*{Acknowledgements}
We thank the Lazouni Clinic (Tlemcen, Algeria) for providing access to Orbscan 3 examinations, supporting data acquisition, and annotation for this study.

% -------------------------------
\section*{Declaration of Competing Interest}
The authors declare that they have no known competing financial interests or personal relationships that could have appeared to influence the work reported in this paper.

% -------------------------------

\section*{Declaration of generative AI and AI-assisted technologies in the manuscript preparation process}
During the preparation of this work the authors used ChatGPT in order to improve language and readability. After using this tool, the authors reviewed and edited the content as needed and take full responsibility for the content of the published article.

%%Harvard
\bibliographystyle{cas-model2-names}
\bibliography{refs}

\end{document}